\documentclass{article}

\usepackage[final,nonatbib]{timeseries_workshop}

\usepackage[utf8]{inputenc} 
\usepackage[T1]{fontenc}    
\usepackage{url}            
\usepackage{booktabs}       
\usepackage{amsfonts}       
\usepackage{nicefrac}       
\usepackage{microtype}      
\usepackage{xcolor}         

\usepackage[pagebackref=true]{hyperref} 
\renewcommand*\backref[1]{\ifx#1\relax \else (Cited on page #1) \fi}

\usepackage{graphicx}
\usepackage{subcaption} 
\usepackage{float}

\usepackage{xcolor}
\definecolor[named]{ACMBlue}{cmyk}{1,0.1,0,0.1}
\definecolor[named]{ACMYellow}{cmyk}{0,0.16,1,0}
\definecolor[named]{ACMOrange}{cmyk}{0,0.42,1,0.01}
\definecolor[named]{ACMRed}{cmyk}{0,0.90,0.86,0}
\definecolor[named]{ACMLightBlue}{cmyk}{0.49,0.01,0,0}
\definecolor[named]{ACMGreen}{cmyk}{0.20,0,1,0.19}
\definecolor[named]{ACMPurple}{cmyk}{0.55,1,0,0.15}
\definecolor[named]{ACMDarkBlue}{cmyk}{1,0.58,0,0.21}
\hypersetup{colorlinks,
  linkcolor=ACMDarkBlue,  
  citecolor=ACMDarkBlue,  
  urlcolor=ACMDarkBlue,   
  filecolor=ACMDarkBlue}

\title{
LiMTR: Time Series Motion Prediction for Diverse Road Users through Multimodal Feature Integration
}

\author{%
  Camiel Oerlemans$^1$\textsuperscript{*}, Bram Grooten$^1$\textsuperscript{*}, Michiel Braat$^{2}$, Alaa Alassi$^{2}$,\\ \textbf{Emilia Silvas}$^{1, 2}$, \textbf{Decebal Constantin Mocanu}$^{3,1}$ \\
  $^1$Eindhoven University of Technology, $^2$TNO Helmond, $^3$University of Luxembourg\\
}

\begin{document}

\renewcommand{\thefootnote}{\fnsymbol{footnote}}
\footnotetext[1]{Equal contribution. Corresponding author: b.grooten@tue.nl}.
\renewcommand{\thefootnote}{\arabic{footnote}}
\setcounter{footnote}{0}

\maketitle

\begin{abstract}
Predicting the behavior of road users accurately is crucial to enable the safe operation of autonomous vehicles in urban or densely populated areas. Therefore, there has been a growing interest in time series motion prediction research, leading to significant advancements in state-of-the-art techniques in recent years. However, the potential of using LiDAR data to capture more detailed local features, such as a person's gaze or posture, remains largely unexplored. To address this, we develop a novel multimodal approach for motion prediction based on the PointNet foundation model architecture, incorporating local LiDAR features. 
Evaluation on the Waymo Open Dataset shows a performance improvement of 6.20\% and 1.58\% in minADE and mAP respectively, when integrated and compared with the previous state-of-the-art MTR. 
We open-source the code of our LiMTR model.\footnote{See \url{https://github.com/Cing2/LiMTR}}
\end{abstract}

\section{Introduction}

Time series motion prediction involves generating potential future trajectories of a road user and estimating their likelihood of occurrence based on a limited set of temporal points \cite{Varadarajan2022}. This is a crucial task in the safe operation of autonomous vehicles, as part of the high-level functional pipeline of autonomous vehicles: \textit{environment perception}, \textit{motion prediction}, \textit{planning}, and \textit{control} \cite{su152014716}. 
Current motion prediction solutions mainly receive a relatively coarse-grained modality; the output of the \textit{object detection} step, which contains the objects' positions, velocities, accelerations, and bounding boxes in combination with accurate road information \cite{Varadarajan2022, shi2024}. This representation of the objects is quite efficient due to their high information density \cite{Gao2020}. However, it potentially leaves out more fine-grained information about a target such as a person's gaze or pose direction \cite{10160609, Chen2023} during the environment perception step.  
Including these modalities could potentially help the model predictions for vulnerable road users, e.g. pedestrians and cyclists. Li et al. \cite{Li2023} have shown that light detection and ranging (LiDAR) data could effectively be used to predict pedestrian crossing. 
We aim to investigate how directly incorporating LiDAR data into motion prediction models can enhance the accuracy of predicted future trajectories, particularly for vulnerable road users.

We introduce a new LiDAR encoder for motion prediction called LiMTR,\footnote{For LiDAR Motion Transformer, pronounced as \textit{Limiter}.} based on the Motion Transformer by Shi et al. \cite[MTR]{shi2022motion}.
We provide the encoder with LiDAR data pertaining only to the target road user, to help the model focus on features such as a person's pose or gaze direction. We refer to this as providing \textit{local} LiDAR features, compared to global scene-level features. 
We demonstrate its effectiveness by experimentation on the Waymo Open Dataset \cite[WOD, see \autoref{app:wod}]{Chen2023}, gaining an overall performance improvement of 6.20\% and 1.58\% in minimum average displacement error (minADE) and mean average precision (mAP) respectively, compared to the baseline without LiDAR.

\section{Time Series Motion Prediction}\label{sec:motion_prediction}

In this section, we explain the task of time series motion prediction, specifically how it is defined for the Waymo Motion Prediction challenge. For this task, the model is provided with 1 second of past information on a scene at 10Hz, and needs to generate up to $m$ potential future trajectories for each of the $n$ road users, referred to as agents, where $m=6$ and $1\leq n \leq 8$ for a single scene. A trajectory consists of the $(x,y)$ coordinates on a local map for 8 seconds into the future at 2Hz. For further background on motion prediction, refer to \autoref{app:background}.

\subsection{Performance Metrics}

We compare the models with metrics used for the Waymo Motion Prediction challenge, also commonly employed in research; for full definitions see \cite{Ettinger2021}. All metrics are computed at time horizons of 3, 5, and 8 seconds and for each class of road user (pedestrian, cyclist, vehicle). Lastly, the mean over these time horizons and classes is taken to compute the final metric.

\textbf{minADE}$\downarrow$ $\in [0, \infty)$ Minimum Average Displacement Error, takes the minimum of the L2-norm between the ground truth trajectory and the $m$ predicted trajectories.

\textbf{MR}$\downarrow$ $\in [0, 1]$ Miss rate is the fraction of trajectory sets where none of the trajectories are correct. A trajectory is determined \textit{correct} if its final position is within a certain threshold distance of the ground truth. This distance threshold is different for lateral and longitudinal distances and scales with the velocity of the target at time 0. 

\textbf{mAP}$\uparrow $ $\in [0, 1]$ Mean Average Precision uses the same definition of correctness as the miss rate to determine true positives and false positives, and only one true positive is allowed per target. The predicted trajectories per target are bucketed over 8 behaviors, i.e. left turn, right turn, etc. Then the area under the precision-recall curve is calculated across various thresholds to get the average precision per bucket. The mean value of these buckets is the mAP.

\section{LiMTR}\label{sec:lidar-mtr}

We proceed to design our LiDAR encoder for motion prediction.
We provide our network with local sections of the raw point cloud directly, training all of our components end-to-end.

\begin{figure*}[]
    \centering
    \includegraphics[width=.9\textwidth]{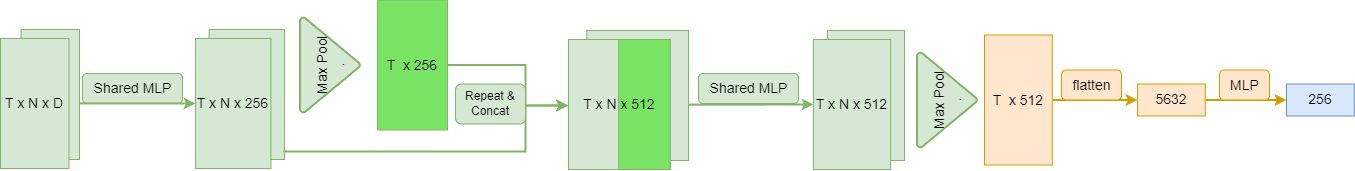}
    \caption{Our LiDAR encoder. Point compression in green, time in orange, with the final feature in blue. The variables represent timesteps (T), points (N), and feature dimension per point (D).}
    \label{fig:lidar_encoder}
\end{figure*}

\subsection{Local LiDAR Features}\label{sec:lidar-preprocess}

The input to our LiDAR encoder is the subset of 3D LiDAR points of each road user we are currently predicting. We select all 3D LiDAR points that fall into the provided bounding box of the object, where we include a 15\% margin, to include possible missed points due to errors in bounding box labeling. We use this subset to greatly reduce computational requirements, and we hypothesize that the LiDAR points of the target are most informative for its future trajectory. This may include potential features such as a person's gaze direction or a cyclist signaling to change course by pointing.

The 3D LiDAR points are centered on the target and rotated to align with the forward-facing direction of the target, similar to the preprocessing of the trajectories \cite{pmlr-v100-chai20a, shi2022motion}. The number of LiDAR points taken is limited to 512; if the number of available points exceeds this, a random subset is taken, if fewer are available the set is padded with zeros. The PointNet architecture \cite{Qi2017} has been shown to be quite robust to random removal of points. Furthermore, we add a one-hot encoding of the target class to each LiDAR point, similar to how Vora et al. \cite{Vora2020} add the probabilities of another detector. Additionally, we include the intensity feature and leave out range and elongation, see \autoref{sec:ablation}.

\subsection{LiDAR Encoder}\label{sec:lidar-encoder}

\autoref{fig:lidar_encoder} presents the architecture of our LiDAR encoder, based on the PointNet \cite{Qi2017} architecture, which was chosen as it can directly process LiDAR points without the need for voxelization or other extensive preprocessing. The diagram shows the dimensions of how a single-object LiDAR point set is transformed into a feature vector; $T=11$ for the number of frames and $N=512$ for the number of points. The encoder consists of two parts: the first part compresses the point dimension and the second compresses the time dimension, resulting in a single feature vector of 256, highlighted in green, orange, and blue, respectively. 
The point compression part uses shared MLPs over the feature dimension and a max-pool layer. Max-pooling is permutation invariant, which is advantageous as the LiDAR points do not have a distinct order \cite{Qi2017}. Additionally, as done in \cite{Qi2017, shi2022motion}, we add a global feature by taking the max-pool, repeating the feature, and concatenating it to the feature dimension. The time compression part flattens the time dimension and then applies multiple MLP layers. 

\begin{figure*}[]
    \centering
    \includegraphics[width=\textwidth]{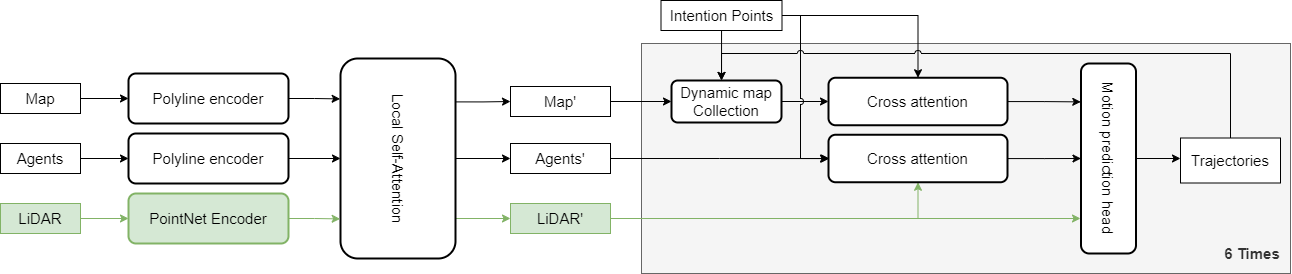}
    \caption{LiMTR architecture based on the MTR \cite{shi2022motion} model, with our LiDAR encoder in green.}
    \label{fig:mtr_lidar}
\end{figure*}

The three MLP blocks consist of 12 linear layers, each followed by Batch Normalization \cite{batch_norm} and ReLU activation \cite{Hahnioser2000}. 
The hidden dimensions within each MLP block are $[256, 512, 1024]$.
From scaling experiments, discussed in \autoref{sec:scaling}, we determined that 12 layers for each shared MLP gave optimal performance. \autoref{fig:mtr_lidar} shows how the LiDAR encoder is integrated into MTR \cite{shi2022motion}. The LiDAR data is first passed through the encoder and then sent to the local self-attention module. Next, the LiDAR feature is combined with the processed agents' features to be used in cross-attention and is also directly sent to the motion prediction head.

\section{Results and Discussion}\label{sec:results}

\subsection{Experimental setup}

We train LiMTR with the AdamW \cite{Loshchilov2019} optimizer with $\beta_1=0.9$, $\beta_2=0.999$ a learning rate of $3\cdot 10^{-4}$, and $0.01$ weight decay. 
We use a Linear Decay learning rate scheduler with a linear warmup period of 5\%, following \cite{Defazio2023}. We train the model for 60 epochs and a batch size of 192 scenarios on 12 NVIDIA A100 GPUs, which takes approximately 80 hours. 
We compare our LiMTR model against multiple baselines, the most important one being MTR \cite{shi2022motion} without the LiDAR modality. Furthermore, we compare against MGTR \cite{Gan2023} which uses global scene-level features, and against Wayformer \cite{Chen2023}, that uses a pretrained object detection model for LiDAR.

\subsection{Main Results}

\autoref{tab:results} presents our results on the Waymo Motion Prediction \cite{Chen2023} validation set. The bottom section shows the performance of Wayformer with \cite{Chen2023} and without \cite{10160609} LiDAR, for which the authors only reported at $t=8$s. The top section shows averages over $t=3$, $5$, $8$s, as is standard practice \cite{Gan2023, shi2022motion}.
Our LiDAR approach shows consistent performance improvement on all metrics compared to MTR with an average increase of 6.20\% and 1.58\% in minADE and mAP, respectively. In comparison to Wayformer with a pretrained LiDAR model, LiMTR achieves an improved mAP, but a slightly higher minADE. 
Their LiDAR solution predicts all objects in the scene which might explain the extra benefit on that metric.
LiMTR performs especially well on vulnerable road users, such as pedestrians and cyclists. These show the largest increase in mAP, suggesting that gazes or poses may indeed be inferred from the LiDAR data.

\begin{table}[]
    \centering
    \caption{Performance on the Waymo Open Dataset \cite{Chen2023}. The \textbf{top} section results are averaged over $t=3$, $5$, and $8$ seconds, the \textbf{bottom} section are reported at $t=8$s. Results of Wayformer, Wayformer+LiDAR, and MGTR are taken from \cite{10160609, Chen2023, Gan2023}. For MGTR \cite{Gan2023}, no minADE or MR metrics are reported. Best results per section are in shown \textbf{bold}.}
    \resizebox{\columnwidth}{!}{%
    \begin{tabular}{llllllllllll}
    \toprule
        & \multicolumn{2}{c}{Avg} & \multicolumn{3}{c}{Vehicle} & \multicolumn{3}{c}{Pedestrian} & \multicolumn{3}{c}{Cyclist}\\
        \cmidrule(lr){2-3} \cmidrule(lr){4-6} \cmidrule(lr){7-9}  \cmidrule(lr){10-12}
        Model & mAP$\uparrow$ & minADE$\downarrow$ & mAP$\uparrow$ & minADE$\downarrow$ & MR$\downarrow$ & mAP$\uparrow$ & minADE$\downarrow$ & MR$\downarrow$ & mAP$\uparrow$ & minADE$\downarrow$ & MR$\downarrow$  \\
    \midrule
        MGTR \cite{Gan2023}             & \textbf{0.45}   &        & \textbf{0.46}  &   &   & \textbf{0.47}  &   &   & \textbf{0.40}  &   &   \\
        MTR \cite{shi2022motion}        & 0.4029 & 0.6812 & 0.4165  & 0.8845  & 0.1756  & 0.4292  & 0.3693  & 0.0855  & 0.3630  & 0.7896  & 0.2097 \\
        LiMTR (Ours)                    & 0.4093 & \textbf{0.6389} & 0.4214  & \textbf{0.7990}  & \textbf{0.1686}  & 0.4363  & \textbf{0.3613}  & \textbf{0.0810}  & 0.3702  & \textbf{0.7565}  & \textbf{0.2041} \\      
    \midrule
        Wayformer \cite{10160609}       & 0.33 & 0.91 & 0.35  & 1.10  & 0.18  & 0.35  & \textbf{0.54}  & 0.11  & 0.29  & 1.08  & 0.22  \\
        Wayformer+LiDAR \cite{Chen2023} & 0.34 & \textbf{0.90} & \textbf{0.37}  & \textbf{1.09}  & \textbf{0.17}  & 0.37  & \textbf{0.54}  & \textbf{0.10}  & 0.28  & \textbf{1.06}  & \textbf{0.21}  \\
        MTR \cite{shi2022motion}        & 0.3459 & 1.1313 & 0.3426  & 1.5151  & 0.2254  & 0.3826  & 0.5985  & 0.1093  & 0.3125  & 1.2802  & 0.2235 \\    
        LiMTR (Ours)                    & \textbf{0.3508} & 1.0496 & 0.3458  & 1.3640  & 0.2156  & \textbf{0.3894}  & 0.5838  & \textbf{0.1002}  & \textbf{0.3171}  & 1.2009  & \textbf{0.2148} \\        
    \bottomrule
    \end{tabular}
    }
    \label{tab:results}
\end{table}

\subsection{Scaling experiments}\label{sec:scaling}

We conduct a network scaling experiment to investigate the LiDAR encoder size required to extract relevant features.
We increase the depth of the 3 MLP blocks simultaneously from 2 to 14 layers with a step of 2. The resulting model ranges from 7.8M to 24M parameters. In the scaling and ablation experiments, we train for $30$ epochs on $10\%$ of the data, using batch size $64$ and learning rate $10^{-4}$. 
\autoref{fig:lidar_scaling} shows the LiDAR encoder size in the number of parameters to minADE.  
The plot indicates that model performance correlates with size, where larger models are preferred. For the main model, we use the 12-layer, 22M parameter model as it demonstrates better performance on the mAP metric.

\subsection{Ablation study}\label{sec:ablation}

We perform ablation experiments to determine two aspects: which LiDAR input features are important and what is the effect of using multiple LiDAR frames. First, as described in \autoref{app:wod}, the dataset also includes additional LiDAR features; range, intensity, and elongation. We trained our model with each feature alone and with all three included. Second, over the 1 second of past data, we have 11 timeframes of LiDAR. To determine the importance of the time series aspect of the LiDAR encoder, we trained LiMTR with 11, 6, 3, or 1 LiDAR timeframe(s) provided.
Each option was trained with 2 different seeds and the mean and standard deviation of the minADE and mAP are shown in \autoref{tab:ablation}.

\paragraph{Time series.} For the number of timestamps, there is no clear distinction on which is better. The full 11 timestamps displays the best performance on mAP, however, the worst performance on minADE. The variance between the runs is too large to determine a clear choice.

\paragraph{Features.} The intensity feature performs better than range and elongation when looking at the minADE metric. Further, the experiments with all features included show similar performance to the intensity feature. This suggests that the intensity feature meaningfully contributes to the task.

\begin{table}[!htb]
    \centering
    \caption{Ablation study on the number of time steps and LiDAR point features.}
    \resizebox{\columnwidth}{!}{%
    \begin{tabular}{ccc|ccc}
    \toprule
    Time steps & mAP $\uparrow$  &  minADE $\downarrow$  & LiDAR features & mAP $\uparrow$  & minADE $\downarrow$ \\
    \midrule
     1   & 0.3333 (0.0005) & 0.9428 (0.0097)             & Range       & 0.3336 (0.0036)           & 0.9608 (0.0001) \\
     3   & 0.3311 (0.0019) & 0.9376 (0.0135)              & Intensity   & 0.3343 (0.0037)           & \textbf{0.9414} (0.0041) \\
     6   & 0.3333 (0.0031)   & \textbf{0.9337} (0.0110)          & Elongation  & 0.3318 (0.0052)           & 0.9734 (0.0066) \\
     11  & \textbf{0.3388} (0.0003)  & 0.9507 (0.0166)        & All        & \textbf{0.3388} (0.0003)  & 0.9507 (0.0166) \\
    \bottomrule
    \end{tabular} 
    }
    \label{tab:ablation}
\end{table}

\section{Conclusion}

We propose LiMTR, a novel time series motion prediction model for autonomous driving, that incorporates local LiDAR features. The model is trained directly on LiDAR data allowing it to capture intricate motion details. When compared to the previous state-of-the-art MTR model, we show that LiMTR gains 6.20\% and 1.58\% in minADE and mAP, respectively. 
By only providing our model with the LiDAR data of a target road user, we help it focus on target-specific features. 
This makes LiMTR complementary to other techniques that incorporate global LiDAR features.

\newpage

\begin{ack}
This work is part of the AMADeuS project of the Open Technology Programme (project number 18489), which is partly financed by the Dutch Research Council (NWO).
This research used the Dutch national e-infrastructure with the support of the SURF Cooperative, using grant number EINF-6221.
\end{ack}

\bibliographystyle{abbrv}
\bibliography{references}

\newpage
\appendix
\section*{Appendix}

\section{Waymo Open Dataset}
\label{app:wod}

The release of the LiDAR data for the Waymo Open Dataset (WOD) \cite{Chen2023} opened up new possibilities for using LiDAR data for motion prediction, which has remained low due to the low availability of large motion prediction datasets with LiDAR data. The Waymo Motion Prediction dataset is two orders of magnitude larger compared to the second largest open motion prediction dataset with LiDAR NuScenes \cite{Caesar2020} with 1k scenarios compared to 104k scenarios for Waymo \cite{Chen2023}. The scenarios in the WOD are often interesting traffic situations, such as an intersection, and the data includes tracked and labeled road users in combination with road layout information.
Each scene is 9 seconds long and recorded at 10 Hz. The data for a scenario consists of three parts: the road information, the agents' information, and LiDAR data. The road and agent information is made with their offboard perception system from the raw sensor data of the car \cite{Ettinger2021}. 

\paragraph{Road information.}
The road information consists of static map features and dynamic features. The latter are traffic light states connected to a lane, given at each timestamp. The static map features represent the 2D road layout and contains a polyline or outline of the following types: lane, road edge, road line, stop sign, crosswalk, speedbump, and driveway.

\paragraph{Agents information.}
The agents' information consists of information on the dynamic objects of the scenario, which are classified as vehicles, cyclists, or pedestrians. Each object consists of a state for every timestamp, where the state has a position in local $x$, $y$, and $z$ coordinates, velocity in $x$ and $y$, and the object's bounding box.

\paragraph{LiDAR.}
The LiDAR data is collected from five LiDAR sensors, one on top of the vehicle with a sensor resolution of 64 pixels in height and 2650 pixels in width, and four on the sides of the vehicle with a sensor resolution of 116x150. Each pixel records its position in $x$, $y$, and $z$ and its corresponding features range, intensity, and elongation.

\section{Motion Prediction Background}
\label{app:background}

In this Appendix, we describe related machine learning work on the problem of time series motion prediction using LiDAR data and explain the MTR \cite{shi2022motion} model used as a baseline for this work. The field of time series motion or trajectory prediction is activity searching for the best and safest algorithms to use for autonomous driving \cite{Chen2023, Gan2023, 10160609, sanchez2023robustness, sanchez2020hybrid}. The starting point for this research is the Waymo Motion Prediction challenge. For this task, the model is given the first second of past data and should predict the exact locations of various road users up to 8 seconds into the future. The provided data consists of road information, road users' information, and raw LiDAR data.

\subsection{Motion Prediction with LiDAR}

There are a few motion prediction works that make use of LiDAR data. Early work by Uber includes LiDAR by voxelizing the data into a birds-eye view (BEV) 3D grid with a resolution of 0.2 \cite{8578474, pmlr-v87-casas18a} 
or 0.16 \cite{10.1109/IV48863.2021.9575718} meters. The voxelized LiDAR is then processed with convolutional neural network (CNN) backbones to extract relevant features and then jointly perform object detection and motion prediction. Similarly, MGTR \cite{Gan2023} also includes BEV 3D voxelized LiDAR data, with two different resolutions of 0.8 and 1.6 meters. These are encoded with a pre-trained LiDARMultiNext \cite{Ye_Zhou_Chen_Xie_Wang_Wang_Foroosh_2023} model and used in their Transformer encoder-decoder architecture to do motion prediction. The MGTR model achieved state-of-the-art performance winning the Waymo Motion Prediction challenge in 2023. 
The voxelization is an important processing step to reduce computation requirements and makes it easier to process with CNN backbones which are good at capturing local structures in the LiDAR data \cite{pmlr-v87-casas18a}. However, the voxelization might leave out finer details due to lowering the resolution, where details that require a higher resolution will not be distinguishable anymore, such as potentially a person's pose or gaze direction.

Waymo, with their release of the 2023 Waymo Open Dataset \cite{Chen2023}, experimented with adding LiDAR data to motion prediction models. They use the final embeddings and detected boxes from a LiDAR 3D object detection model, SWFormer \cite{Sun2022}, and include these in the WayFormer \cite{10160609} model. They lowered the prediction threshold of the object detection model allowing it to detect more objects. Wayformer with the addition of the LiDAR data gained an increase in mAP from 0.35 to 0.37 for both vehicles and pedestrians. However, there was no improvement for cyclists, which may be due to the features their LiDAR encoder retrieves for bounding box detection might leave out information relevant to motion prediction. This suggests a need to investigate if the features extracted by an object detection model are adequate for motion prediction.

\subsection{Motion TRansformer: MTR}

The MTR model \cite{shi2022motion} is the baseline for this work and the model we extend with our LiDAR approach. 
We select MTR as the open-source state-of-the-art during our research, following the first edition of the MTR model \cite{shi2022motion}. 
Shi et al. later published the improved MTR++ \cite{shi2024} version that includes symmetric scene context encoding and joint motion decoder, allowing the model to predict multiple objects in the same scene more efficiently. The MTR model uses the Transformer encoder-decoder architecture. \autoref{fig:mtr_lidar} shows an outline of the model, where the grey parts are the original architecture.

\paragraph{Data representation.}
The road information and agent history are encoded using the vectorized representation \cite{Gao2020}.
The coordinates of the road and agent polylines are centered and rotated on the current target agent coordinate system, adopting an agent-centric strategy \cite{9710037, Varadarajan2022}.

\paragraph{Context encoder.}
First, a polyline encoder based on the PointNet \cite{Qi2017} architecture is used to encode the agent and road vectors to produce their respective features. These encoded features are then processed with local self-attention by only attending to nearby objects. This is done by applying k-nearest neighbors (KNN) on each object based on its initial position to collect a neighborhood of \textit{k} neighbors.
This reduces the computation requirements and crucially preserves the local structure of the scene \cite{shi2024}.

\paragraph{Motion decoder.}
The motion decoder uses the transformer-based decoder structure to apply cross-attention on the previously encoded road and agent features. They employ learnable query embeddings from intention points to act as intention queries. The intention points are made by applying k-means clustering on the set of relative final locations of all targets in the training set, separately for the three classes, vehicles, pedestrians, and cyclists. These intention points act as anchor points, similar to \cite{9710037}
goal points, and are crucial to generating a distinct set of trajectories. The query embeddings are then used in cross-attention with the agent features to collect relevant features for each intention point.  

Before applying attention to map features, a \textit{dynamic map collection} technique is used. Similar to the context encoder local attention method, this also applies KNN on the map features to collect a set of relevant features. For the first decoder layer, the distance to the intention point is used for KNN, and in consecutive layers, the smallest distance to any point in the previously predicted trajectories. This allows the model to iteratively refine the trajectory on relevant features. 

The motion prediction head takes the queried features per intention point and uses multilayer perceptron (MLP) layers to generate the trajectories with accompanying probabilities. A trajectory consists of 80 timestamps which cover 8 seconds with 10Hz. Following \cite{pmlr-v100-chai20a, Varadarajan2022} they represent the trajectories as Gaussian Mixture Model (GMM) components $x, y, std_x, std_y, \rho$, for each timestamp. This is an important part of modeling the ambiguous behavior of the agents.

\paragraph{Loss.}
The loss function consists of three parts, the negative log-likelihood loss for the GMM components, an $L1$ loss on the predicted velocity, and a cross-entropy loss on the probabilities of the trajectories.

\section{Analysis}

In this section we include additional figures and information that complement the scaling and ablation studies in the paper.
For the time series ablation: we take equally spaced timestamps over the $11$ total. Thus, e.g., for $6$ time steps these are the frames $(0,2,4,6,8,10)$, and for the $1$ timeframe we use the last time step. 

\begin{figure}
    \centering
    \includegraphics[width=0.9\linewidth]{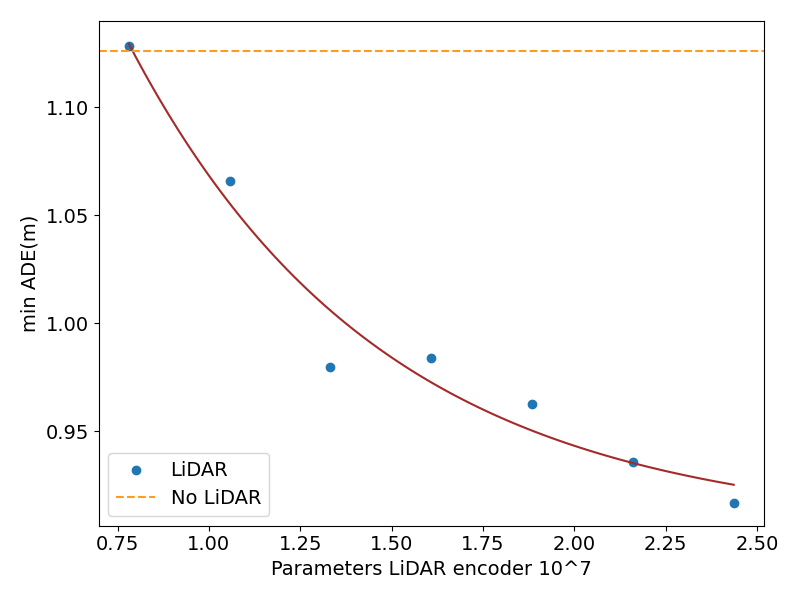}
    \caption{Scaling of the LiDAR encoder network, showing the number of parameters to model performance in minADE, including a horizontal line denoting baseline MTR performance without LiDAR, and a fitted exponential function.}
    \label{fig:lidar_scaling}
\end{figure}

\begin{figure}
\begin{subfigure}{.5\linewidth}
  \centering
  \includegraphics[width=.99\linewidth]{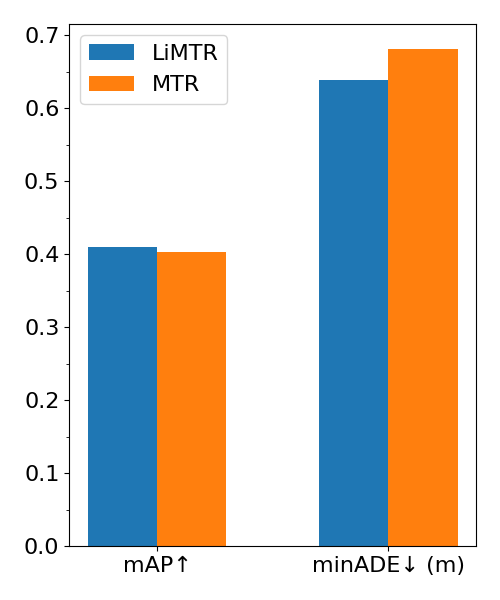}
\end{subfigure}%
\begin{subfigure}{.5\linewidth}
  \centering
  \includegraphics[width=.99\linewidth]{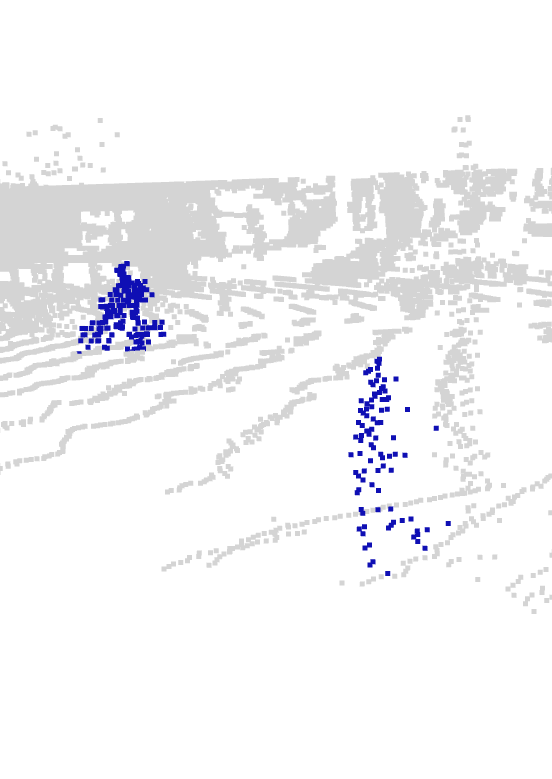}
\end{subfigure}
  \caption{(\textbf{Left}) Performance of LiMTR (ours) and MTR. (\textbf{Right}) Local LiDAR data of a cyclist and a pedestrian in a scene in blue. Our LiMTR model receives time series LiDAR data at 10Hz, capturing the movement of such point clouds per target object.}
    \label{fig:limtr}
\end{figure}

\end{document}